\documentclass[fleqn,10pt]{wlscirep}
\usepackage[utf8]{inputenc}
\usepackage[T1]{fontenc}
\usepackage{lineno}

\usepackage{lineno,hyperref}

\usepackage{enumitem}
\usepackage{amsmath}
\usepackage{enumitem}
\usepackage{tabularx}
\usepackage{booktabs}
\usepackage{comment}
\usepackage{url}
\usepackage{float}
\usepackage{hyperref}
\hypersetup{
    colorlinks=true,
    linkcolor=blue,
    filecolor=magenta,
    urlcolor=cyan,
}
\usepackage{breakurl}

\newcommand{\hide}[1]{}
\newcounter{comments}

\title{An Artificial Intelligence Dataset for Solar Energy Locations in India}

\author[1,$\dag$, *]{Anthony Ortiz}
\author[2,$\dag$]{Dhaval Negandhi}
\author[3]{Sagar R Mysorekar}
\author[3]{Joseph Kiesecker}
\author[3]{Shivaprakash K Nagaraju}
\author[1]{Caleb Robinson}
\author[3]{Priyal Bhatia}
\author[3]{Aditi Khurana}
\author[1]{Jane Wang}
\author[1]{Felipe Oviedo}
\author[1, *]{Juan Lavista Ferres}
\affil[1]{Microsoft AI for Good Research Lab, Redmond, WA, USA}
\affil[3]{Forum for the Future}
\affil[3]{The Nature of Conservancy (TNC)}

\affil[*]{corresponding authors: Anthony Ortiz (anthony.ortiz@microsoft.com), Juan Lavista (jlavista@microsoft.com)}

\affil[$\dag$]{Authors contributed equally to this work}

\begin{abstract}
Rapid development of renewable energy sources, particularly solar photovoltaics (PV), is critical to mitigate climate change. As a result, India has set ambitious goals to install 500 gigawatts of solar energy capacity by 2030. Given the large footprint projected to meet renewables energy targets, the potential for land use conflicts over environmental values is high. To expedite development of solar energy, land use planners will need access to up-to-date and accurate geo-spatial information of PV infrastructure. In this work, we developed a spatially explicit machine learning model to map utility-scale solar projects across India using freely available satellite imagery with a mean accuracy of 92\%. Our model predictions were validated by human experts to obtain a dataset of 1363 solar PV farms. Using this dataset, we measure the solar footprint across India and quantified the degree of landcover modification associated with the development of PV infrastructure. Our analysis indicates that over 74\% of solar development In India was built on landcover types that have natural ecosystem preservation, or agricultural value.

\end{abstract}
\begin{document}

\flushbottom
\maketitle

\thispagestyle{empty}

\section*{Background \& Summary}

\begin{figure*}[!th]
    \centering
    \includegraphics[width=.9\linewidth]{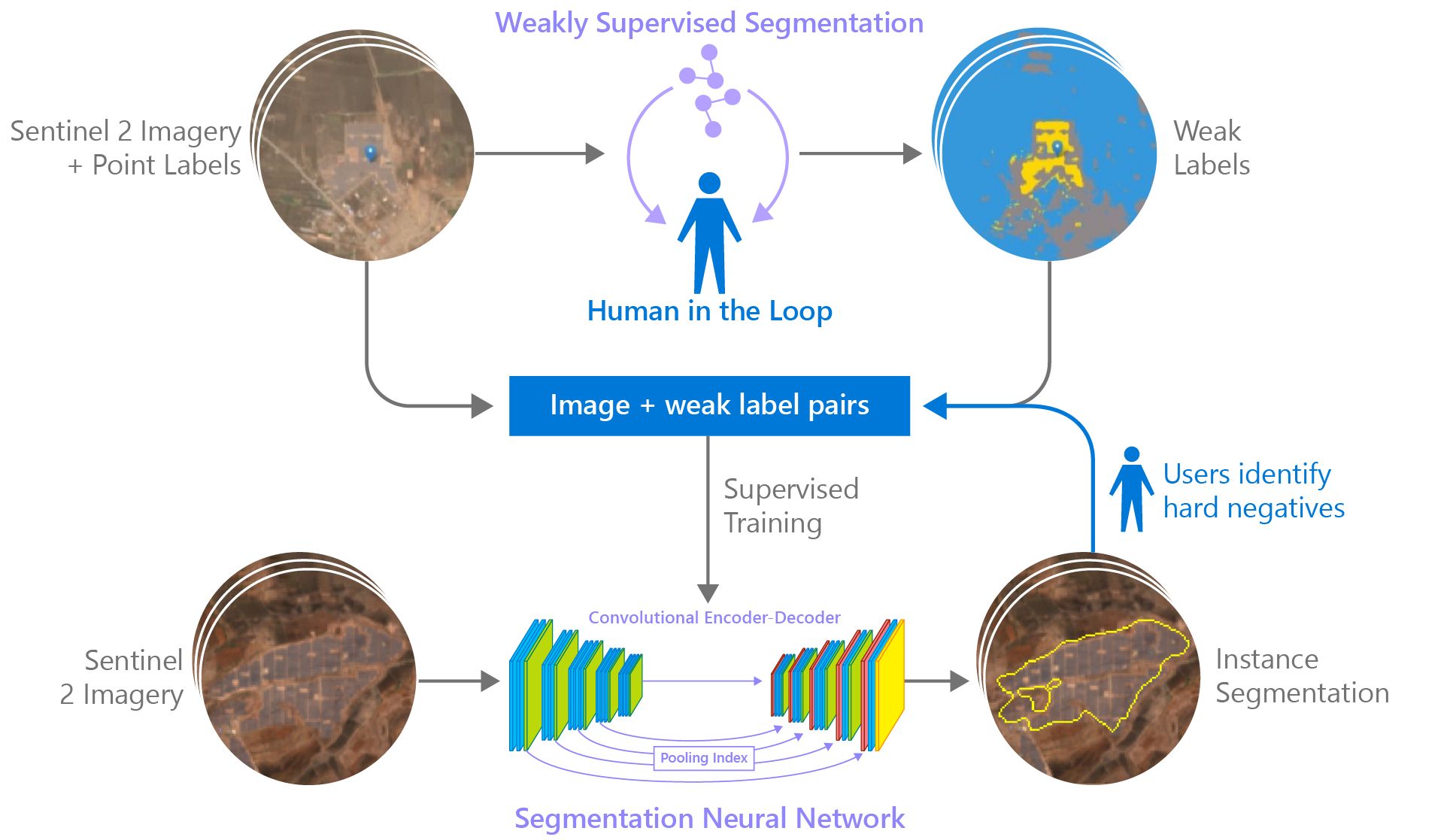}
    \caption{Proposed solar PV mapping pipeline. Given a small set of point labels and its corresponding Sentinel 2 imagery, pixels are clustered into multiple clusters (64 for our experiments). These clusters are merge into a user defined smaller set of classes (three in this example) using a linear classifier. Cluster merge results are shown in a web tool where a human user provides feedback on which pixels belong to the solar farms class or to the other background classes and the linear classifier if finetuned based on the feedback from the user. This weakly supervised segmentation process is represented at the top of this figure and is interactively performed to obtain weak semantic labels like the example shown at the top right of the figure. These labels paired with the corresponding geo-located Sentinel 2 image are used to    create a semantic segmentation dataset suitable for supervised training of a solar farm semantic segmentation model. The obtained segmentation neural network can be used to perform inference for solar farms in novel scenes as shown at the bottom of the figure. False positive predictions are considered hard negatives and are used to augment the training dataset and finetune the supervised segmentation neural network improving its false positive rate. This process of performing inference in novel scenes, adding hard negative to the training set and finetuning the supervised model further can be repeated multiple times until the performance of the results is good enough for large scale inference.}
    \label{fig:pipeline}
\end{figure*}

India is rapidly expanding its deployment of clean energy~\cite{climatescope2020}. The dual benefits of climate mitigation potential, and lower cost of production, makes renewable energy cost-competitive compared to coal and other conventional energy sources. Therefore, to achieve the nationally determined contribution (NDC) targets such as: 40\% share of non-fossil fuel cumulative power generation capacity, and to halt greenhouse gasses (GHGs) emission from fossil fuels, India has committed to 500 gigawatts (GW) of installed renewable energy capacity by 2030~\cite{ceainstallcapacity2021}. India intends to reach 225 GW of renewable power capacity by 2022 exceeding the target of 175 GW pledged during the Paris Agreement. As of 2018 India ranks fifth in installed renewable energy capacity with fourth most attractive renewable energy market in the world.

Solar energy is expected to play an increasingly large role in India’s clean energy transition. Of the 2030 (500 GW) target by 2030, solar energy is expected to contribute 300 GW~\cite{ceaoptimal}. Over the last five years, the installed capacity for solar energy has increased more than five-folds~\cite{ceainstallcapacity2020}. Of the total RE capacity added during this period, more than two-thirds has come from utility-scale solar photovoltaic. Solar energy companies in India project the same trend to continue over the next five years with utility-scale solar energy expected to add 39 GW of the 60 GW of installed RE capacity~\cite{ceosurvey2020}.

Despite the policy commitments in India, many studies have questioned the land-based targets for solar energy deployment and have highlighted the difficulties related to disputes over land use~\cite{mohan2017whose,kiesecker2020renewable}. Renewable energy requires a huge amount of space~\cite{kareiva2017energy}. If these energy installations aren’t sited carefully, they can cause significant damage to wildlife, natural habitats and critical ecosystem services and even generate greenhouse gas emissions that reduce their climate benefits~\cite{kiesecker2019hitting}. Despite the recognition of these challenges policy makers and governments have struggled to maintain robust geospatial information on the rapid expansion of renewable energy technologies. Access to these data will be critical to assess past impacts and planning to avoid future conflicts.

At present, there is limited information that is compiled and publicly available on the location of utility-scale solar photovoltaic projects across the country. Most location information for a project is typically limited to its associated jurisdictional boundary. The lack of more specific information, such as project boundaries, makes it difficult to identify factors that may be driving land suitability for such projects, and thus deprive policy-makers of the relevant information to expedite development. In addition, without such information, it is difficult to understand the nature of land-use changes driven by solar energy in India. This is particularly significant as some land-use changes from solar development (e.g., from biodiversity-rich habitats, or those places that are important for agriculture or pasture lands for local grazing-dependent communities) may lead to socio-ecological land conflicts and ultimately slow the transition to renewable energy. While some of the facility-level location information is collected by government agencies during project-level planning and construction phases, this information is not typically made publicly available. Other datasets that are publicly available, i.e. OpenStreetMap, usually do not capture the full range of development given sampling biases of these crowd sourcing approaches. Fortunately, freely available high-resolution remotely sensed imagery and new artificial intelligence techniques make it possible to now map utility-scale projects~\cite{davidson2014modeling,yu2018deepsolar,malof2016automatic,kruitwagen2021global}. 

We present the first country-wide database of solar photovoltaic farms for the country of India and show that it is feasible to also detect when the solar farms were created – allowing for further land use and sustainable development analysis. Our contributions are twofold:

\begin{enumerate}
    \item A novel methodology for creating datasets of remotely sensed objects using satellite imagery when labeled data available is limited. This new method consists of a semantic segmentation model trained in stages using human-machine interaction and hard negative mining (HNM).
    \item The quantification of land cover change associated with solar energy development in India. This analysis can inform policy makers to develop policies ensuring renewable energy is developed in low conflict areas.
\end{enumerate}

\section*{Materials and Methods}
Datasets are often created using human experts of crowdsourced labelers. However, there are use cases, like detecting small objects on the surface of the earth, where this task is costly, time consuming, and unscalable. When sufficient labeled data is available, machine learning models tend to be helpful reducing the time required to accomplish this task. Here we present a methodology for creating datasets of remotely sensed objects using satellite imagery when labeled data available is limited.  To develop our map of utility-scale solar arrays across India first we assembled point labels of known solar PV farms and used human-machine interaction for a user to finetune an unsupervised model to create weak segmentation labels, labels obtained through weakly supervised learning~\cite{zhou2018brief}, of the solar farms. Then we paired these weak pixel-wise segmentation labels with geo-located Sentinel 2 imagery to train a supervised segmentation neural network and further improved in multiple stages of Hard Negative Mining (HNM). Finally, we estimated when solar PV installations were constructed and assessed the land use prior to construction for each array. Finally human experts validated the output of the AI model and individual solar arrays were clustered into solar farms using distance-based clustering. Figure \ref{fig:pipeline} describes the proposed methodology.  

\paragraph{Solar farms point labels dataset} We used a set of 117 geo-referenced point labels corresponding to the center point of different solar installations for the states of Madhya Pradesh (45-point labels) and Maharashtra (72-point labels) in India to train our initial solar mapping model. We also obtained 191 noisy solar installations point labels for four other Indian states including Kerala (15), Telangana (28), Karnataka (73), Andhra Pradesh (75). The noisy points labels did not accurately match the exact solar installation location. These labels were obtained using previously mapped solar farms through OSM and other Nature of Conservancy (TNC) partners. 

\paragraph{Sentinel 2 (S2) satellite Imagery}

The Sentinel-2 program developed by the European Space Agency (ESA) provides global imagery in thirteen spectral bands at 10m-60m spatial resolution and a revisit time of approximately five days free of cost. In this work, we use 12 of the available spectral bands while excluding S2 Band 10 which is used mostly to mask out clouds since cloudy scenes were filtered out as the input to the solar mapping model.

\paragraph{Copernicus Global Land Cover} The Dynamic Land Cover map at 100 m resolution (CGLS-LC100) from Copernicus provides global land cover map at 100 m spatial resolution for the period 2015-2019 over the entire Globe, derived from the PROBA-V 100 m time-series. The product also includes all basic land cover classes including shrubs, herbaceous vegetation, cultivated and managed vegetation / agriculture, urban / built up, bare / sparse vegetation, snow and ice, permanent water bodies, and more.

\paragraph{NRSC Land Use Land Cover~\cite{lulc2017}} 

\begin{figure}[!th]
    \centering
    \includegraphics[width=.6\linewidth]{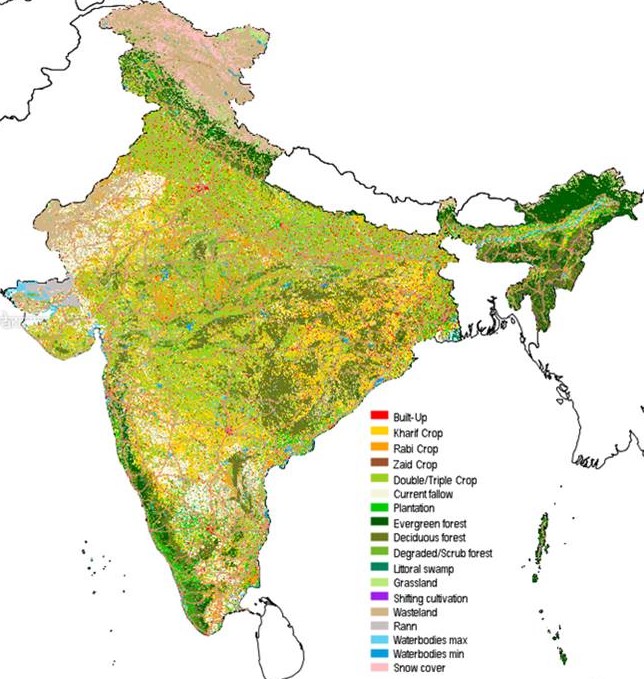}
    \caption{Landcover data at 60 m/px spatial resolution for the country of India for the year 2017. Note: The boundaries of India shown here are neither authenticated nor verified and are not to scale. They are only meant for graphical representation. All efforts have been made to make them accurate, however, neither Microsoft nor TNC own any responsibility for the correctness or authenticity of the same.}
    \label{fig:landcover-india}
\end{figure}

Land Use Land Cover (LULC) maps for the country of India generated by the National Remote Sensing Centre (NRSC) at the Indian Space Research Organization. Annual land use / land cover mapping is carried out at 1:250k scale and is made available at approximately 60m/px resolution.  Figure \ref{fig:landcover-india} shows a snapshot of the Land Use Land Cover data for the year 2017 at a 50m/px resolution along a legend for the classes covered. This data along the Copernicus Global Land Cover is used for the land cover change analysis.

\section*{Semi-supervised label generation: from point labels to semantic annotations}\label{semi-supervision}

\begin{figure*}[!th]
    \centering
    \includegraphics[width=\linewidth]{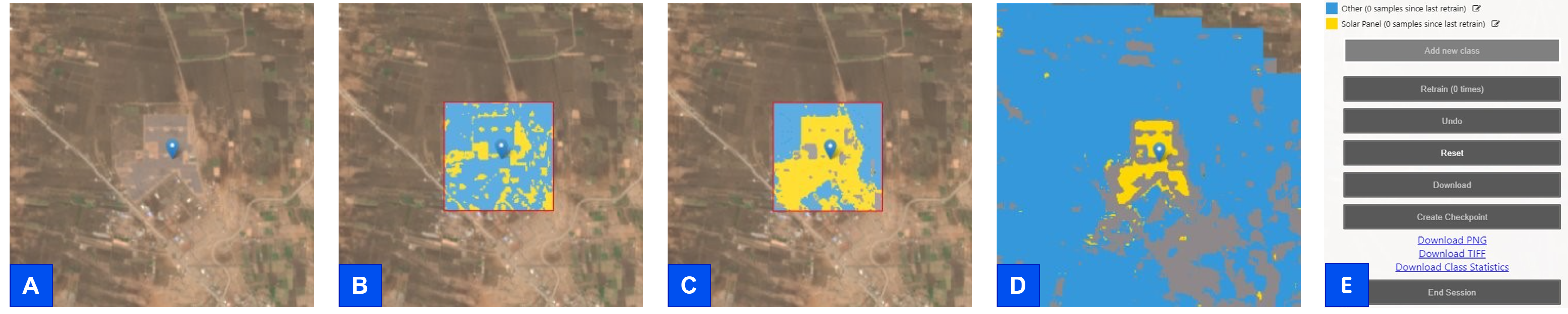}
    \caption{Human-Machine interaction for unsupervised semantic label generation pipeline. (\textbf{A}) Use point labels to find features, (\textbf{B}) Initial unsupervised model will segment imagery at pixel level by color, (\textbf{C}) Fine-tuning to segment solar farms (yellow) vs other (blue, grey), (\textbf{D}) Apply the fine-tuned model to generate weak pixel-wise labels (\textbf{E}) Download labels generated in D as GeoTIFFs to incorporate into a solar installation semantic segmentation dataset of noisy semantic labels.}
    \label{fig:unsupervised}
\end{figure*}

Finding solar installations from satellite imagery can be formulated as a semantic segmentation computer vision task. The goal of semantic image segmentation is to label each pixel of an image with a corresponding class of what is being represented~\cite{long2015fully}. However, pixel-wise labels are required for  semantic segmentation~\cite{ortiz2018integrated}. Manually creating segmentation labels is costly and time consuming. This problem exacerbates while working with noisy point labels with non-systematic displacements errors as it is the case for the solar farms point label dataset described earlier. To overcome this limitation, and generate semantic labels at scale we first pre-trained a convolutional neural network to cluster pixels from Sentinel 2 satellite imagery by color in an unsupervised manner. We, then, used an interactive web application similar to the one proposed by Robinson et al.~\cite{robinson2020human} to quickly fine-tune the network to cluster pixels corresponding to solar installations into a single solar installation class as shown in Figure~\ref{fig:unsupervised}. This fine-tuned model is then used to obtain noisy/weak semantic labels for all available solar farm point labels as shown in Figure \ref{fig:unsupervised} parts D and E. The obtained pixel-wise labels make it possible to create a small semantic segmentation dataset suitable to train supervised semantic segmentation models.

\begin{figure*}[th]
    \centering
    \includegraphics[width=\linewidth]{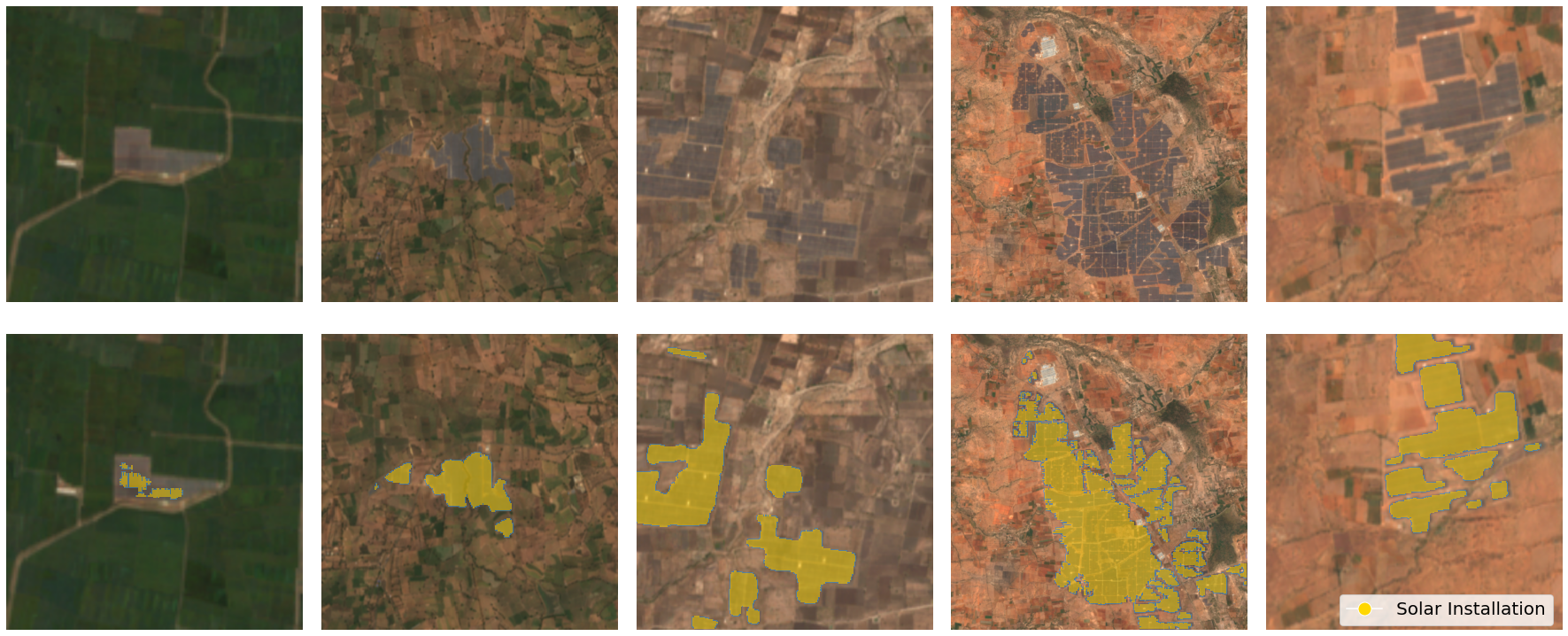}
    \caption{Example predictions. Examples of correct solar farm predictions from our model for different areas in India. Our model accurately outlines solar farms across different areas and backgrounds.}
    \label{fig:solarpredictions}
\end{figure*}
\paragraph{Weak labels solar PV installations segmentation dataset} Following the described semi-supervised semantic label generation approach applied to the solar farms point labels dataset for all states but Maharashtra, we generated an initial segmentation dataset consisting of $234$ pairs of Sentinel 2 image patches of size $256 \times 256$ pixels containing solar PV installations and corresponding pixel-wise labels for the classes ``background'' (0) and ``solar PV installation'' (1) and $50$ pairs of randomly sampled images patches without solar installations with the corresponding pixel-wise labels. The dataset was split into training (80\%), validation (10\%), and test (10\%) disjoint sets. 

\paragraph{Pristine labels solar PV farms test set}
The 72 locations with known solar farms from the point label dataset from Maharashtra, we manually labeled the outlines of the solar farms. These polygons along with corresponding Sentinel 2 imagery constitute what we call the pristine labels solar PV farms and were reserved for testing the models.

\subsection*{Supervised Semantic Segmentation of Solar Farms}\label{supervised}

 Now we formalize our solar farms mapping approach. Let $\left(x_{n}\right)_{n = 1}^{N}$ represent a set of training Sentinel 2 satellite image patches. Each image patch $x_n$ is associated with a corresponding pixel-wise semantic segmentation mask. For each pixel $(i,j)$ in the image patch $x_n$ we aim to assign a label $l_n = 1$ when the pixel belongs to a solar installation and  $l_n = 0$ otherwise. For the segmentation of solar installations we trained several U-Net models~\cite{ronneberger2015u} with different depths and number of input filters on the solar PV installations segmentation training set. We used the Adam optimizer~\cite{kingma2014adam} with a batch size of 32 to train all our models. All neural network models were trained from randomly initialized weights using a learning rate (LR) of 0.001 (The LR hyperparameter controls how much the model weights change in response to the estimated error each time the model weights are updated) for 50 epochs (i.e., we showed the neural network all training samples 50 times). We decay the learning rate by 10\% after 5 epochs of no performance improvement in the validation set. Weighted binary cross-entropy was used as the loss function. The model architecture with best performance in the validation set was selected for the rest of the experiments. 
 
 \paragraph{Hard Negative Mining (HNM)} The previously described dataset contains ``easy'' background examples obtaining from a random sampling procedure.  Models trained on the created dataset will see far more ``easy'' negative samples from background regions than difficult negative samples from areas similar in appearance, shape, or spectral signature solar PV installations. It has been shown that some form of hard negative mining is useful to improve the performance of object detectors~\cite{jin2018unsupervised,zhao2019object}. In this work, we adopt a bootstrapping~\cite{wan2016bootstrapping} approach where we train an initial model and test it by doing inference across different new sentinel image tiles. Inference results were visually inspected for false positive predictions. These false positive predictions represent ``hard negative samples'' and were added to the train set of the solar PV segmentation dataset. The segmentation model can now be re-trained using the new training set for better performance. The HNM procedure can be repeated multiple times.

\paragraph{Predictions post-processing} To reduce the number of false positive identifications still predicted by the model, we incorporated OpenStreetMap~\cite{OpenStreetMap} data to remove false positive predictions over road areas. We also used the Normalized Difference Snow Index (NDSI)~\cite{hall2010normalized} and the Normalized Difference Water Index (NDWI)~\cite{gao1996ndwi} to remove false positive predictions around snow and water bodies respectively.

\subsection*{Solar farms initial development }\label{tcm}

\begin{figure*}[!th]
\centering
\includegraphics[width=1\linewidth]{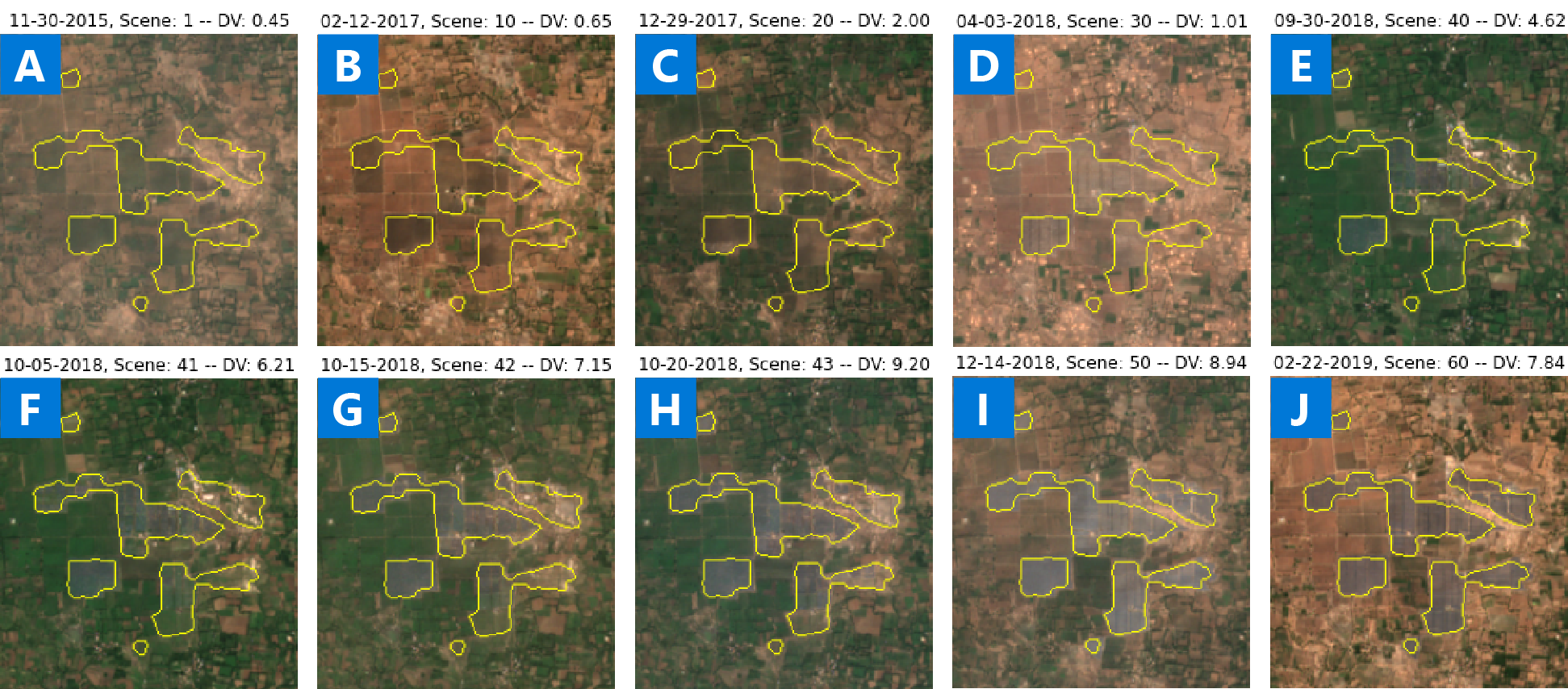}
\caption{Set of sample scenes of cloud free Sentinel 2 imagery from time series for a predicted solar farm used as input to the Temporal Cluster Matching algorithm.}
\label{fig:solar-time-series}
\end{figure*}

\begin{figure*}[!th]
\centering
\includegraphics[width=.6\linewidth]{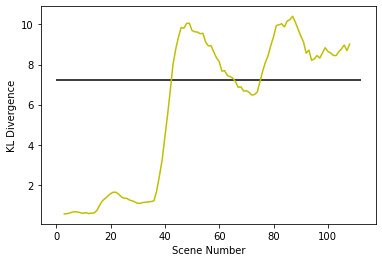}
\caption{KL Divergence TCM corresponding to the time series imagery used on Figure~\ref{fig:solar-time-series}. TCM successfully predict scene 41 as the scene in which the initial development of the solar farm is first observed.}
\label{fig:kl-tcm}
\end{figure*}

We use Microsoft's Planetary Computer~\footnote{Microsoft’s Planetary Computer is freely available at: \url{https://planetarycomputer.microsoft.com/}} to query all available Sentinel 2 cloud free imagery between 2015 and December of 2020 matching the outline of each of the predicted solar PV farms.  We apply Temporal Cluster Matching (TCM), an algorithm for detecting changes in time series of remotely sensed imagery when footprint labels are only available for a single point in time~\cite{robinson2021temporal}, to the Sentinel 2 Imagery time series obtain from the planetary computer to identify when the detected solar farms from 2020 were first built using Sentinel 2 temporal. Figure \ref{fig:kl-tcm} shows the KL divergence for all scenes in the S2 imagery time series used as input for the solar farm shown in Figure \ref{fig:solar-time-series}. The black horizontal line represents the median of the KL divergence values. The median KL divergence is used as threshold to determine the scene of initial development. TCM successfully predict scene 41 as the scene in which the initial development of the solar farm is first observed. The estimated year of development is included for each solar farm in the released dataset. Scene 41 along with scenes pre and post development are shown on Figure \ref{fig:solar-time-series} along with the dates the scene was collected and the TCM computed KL divergence values. The estimated year of development is included for each solar farm in the released dataset.

\subsection*{Landcover change analysis}

\begin{figure}[!th]
\centering
\includegraphics[width=.6\linewidth]{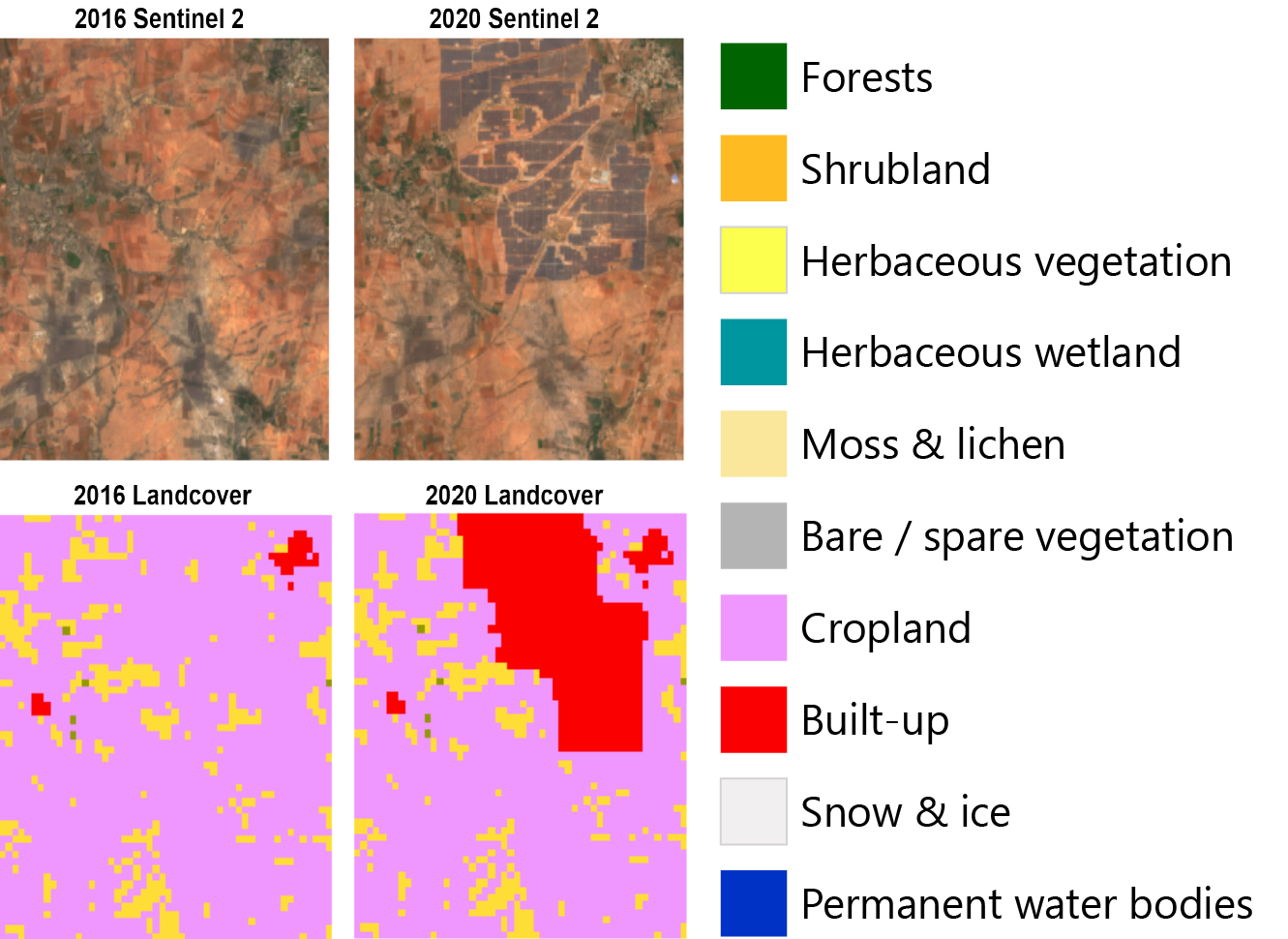}
\caption{Sentinel 2 imagery before the detected solar installation was built (2016) and after it was built (2020) along with the corresponding landcover low resolution information. This information can be used to estimate how landcover changed to support the building of solar installations at scale. Note that solar PV installation is not a land cover type covered on this dataset.}
\label{fig:landcover}
\end{figure}

The year of initial development obtained using TCM, along with the Copernicus annual Global Land Cover from 2015 to 2019 and the NRSC Land Use Land Cover data from 2017 previously described, facilitates the study on environmental and socio-economic implications of solar photovoltaic energy development by analyzing which landcover classes are being impacted by solar farms. Figure~\ref{fig:landcover} shows Sentinel 2 imagery before the detected solar installation was built (2016) and after it was built (2020) along with the corresponding landcover from Copernicus annual Global Land Cover to illustrate how it can be informative of the type of landcover being impacted by the installation of the solar PV farms.

\begin{table*}[ht]
\caption{Performance of proposed model in held out test set of weak labels solar PV segmentation dataset}
\label{table:segmentation-performance}
\centering
\begin{tabular}{@{} l  r r r r @{}}
\toprule 
 \multicolumn{5}{c}{\textbf{Large Noisy Labels Test Set}}  \\
        \midrule
\textbf{Model} & \multicolumn{1}{c}{\textbf{IoU (\%)}} & \multicolumn{1}{c}{\textbf{Mean Acc (\%)}} & \multicolumn{1}{c}{\textbf{Pix Recall (\%)}} & \multicolumn{1}{c}{\textbf{Pix Precision (\%)}} \\ \midrule
U-Net Model   &  59.79   & 85.81  & \textbf{75.80} &  73.67 \\ 
U-Net Model + HNM    & 59.52   &   86.39 &  73.45  & 75.08 \\ 
U-Net Model + 2HNM   & 60.29 &   89.03  &  70.88   &  78.35  \\ 
Model + 2HNM + Post  & \textbf{68.87} &   \textbf{94.76}  & 70.72 &      \textbf{84.63}  \\ 
\bottomrule
\end{tabular}
\end{table*}

\begin{table*}[!tb]
\caption{Performance of proposed model in held out test set of pristine labels}
\label{table:segmentation-performance-pristine}
\centering
\begin{tabular}{ l  r r r r r }
\toprule
 \multicolumn{6}{c}{\textbf{Pristine Labels Test Set}}  \\
        \midrule
\textbf{Model} & \multicolumn{1}{c}{\textbf{IoU (\%)}} & \multicolumn{1}{c}{\textbf{Mean Acc (\%)}} & \multicolumn{1}{c}{\textbf{Pix Recall (\%)}} & \multicolumn{1}{c}{\textbf{Pix Precision (\%)}} &
\multicolumn{1}{c}{\textbf{Recall (\%)}}\\ 
\midrule
U-Net Model + 2HNM   & 80.67 &   95.62   &  86.59  & 91.03 & 94.4  \\ 
\bottomrule
\end{tabular}
\end{table*}

\section*{Data Records}
Our solar farms dataset is stored in vector data form for the use of the community. The final dataset includes 1363 validated and grouped solar PV installations. We provide the vector data in the form of polygons or multi-polygons outlining the solar farms and center points with the geo-coordinates for the center of each installation in a file named: \emph{\url{https://github.com/microsoft/solar-farms-mapping/blob/main/data/solar_farms_india_2021.geojson}}.

The dataset includes the following variables:
\begin{itemize}
\item fid: Unique identifier for a solar farm. It is shared by polygons belonging to the same farm.
\item Area: The area of the site in squared meters ($m^2$).
\item Latitude: Latitude corresponding to the center point of the solar installation.
\item Longitude: Longitude corresponding to the center point of the solar installation.
\item State: Indian State where the solar PV installation is located at.
\end{itemize}
The raw data format can also be found at Zenodo at: \url{https://zenodo.org/record/5842519#.Yn7UT_iZND9} under Creative Commons Attribution 4.0 International license~\cite{ortiz2022artificial}

\section*{Technical Validation}

Table \ref{table:segmentation-performance} shows the performance of our model using the test set from our Solar PV installations segmentation dataset described in the previous section. We can observe how pixel-wise intersection over the union (IoU), mean pixel accuracy, and pixel-wise precision (rate of correct pixel predictions among all positive pixel predictions) improve by retraining the model with hard negative samples while pixel-wise recall (rate of correct pixel predictions among all pixels corresponding to a solar PV installation) decreases slightly as hard negative mining forces the model to be more conservative. Object-wise metrics like farm-wise recall (rate of correct solar-farm detections among all solar PV farms) better describe the performance of the model for this task since missing certain solar farm pixels have no practical effect in being able to detect the solar installation. Pixel-wise metrics are also more susceptible to being affected by the noisy nature of the dataset.

Table \ref{table:segmentation-performance-pristine} shows the performance of our model in the small pristine solar PV test set of manually labeled solar farms. Our best model shows 80.7\% intersection over the union and mean pixel accuracy of 95.6\%, a pixel-wise precision of 91\%, a pixel-wise recall of 86.6\% and a farm-wise Recall of $94.4\%$ before post-processing. This indicates that lower pixel-wise performance in the weak labels solar PV installations dataset might be due to the noise in the ground truth. 

For qualitative results, Figure \ref{fig:solarpredictions} shows sample predictions from our model for a diverse set of image patches at different scales and under different background conditions. It shows robust segmentation performance across different locations. Figure~\ref{fig:solar-time-series} shows predictions over time for a single solar farm. It shows consistent performance in different imaging conditions.

\subsection*{Pearson Correlation Analysis of Historical Solar Install Capacity  and Temporal Model Predictions}\label{s2-temporal}


\begin{table}[ht]
\caption{Karnataka state solar installation capacity}
\label{table:install-capacity}
\centering
\begin{tabular}{| l | c |}
\hline
\textbf{Time} & {\textbf{Solar Installation Capacity (MBH)}} \\ \hline
March 2016  & 147 \\ 
\hline
March 2017  & 1039 \\ 
\hline
March 2018  & 4960 \\ 
\hline
March 2019  & 5944 \\ 
\hline
March 2020  & 7046 \\ 
\hline
\end{tabular}
\end{table}

The state of Karnataka in India provides information about the installed solar photovoltaic installed capacity since 2016 as shown in table \ref{table:install-capacity}. We obtain annual Sentinel 2 median composites using all available scenes with under 3\% cloud coverage obtained between January and May for the years 2016-2020 covering the entire state of Karnataka. For the years 2016, 2017, and 2018 the surface reflectance Sentinel products were not available. To alleviate covariate shift we perform tile-wise histogram matching~\cite{shapira2013multiple} from Top of Atmosphere (ToA) Sentinel 2 median composites to the 2020 tiles surface reflectance Sentinel 2 median composites. 

\begin{figure*}[!th]
\centering
\includegraphics[width=1\linewidth]{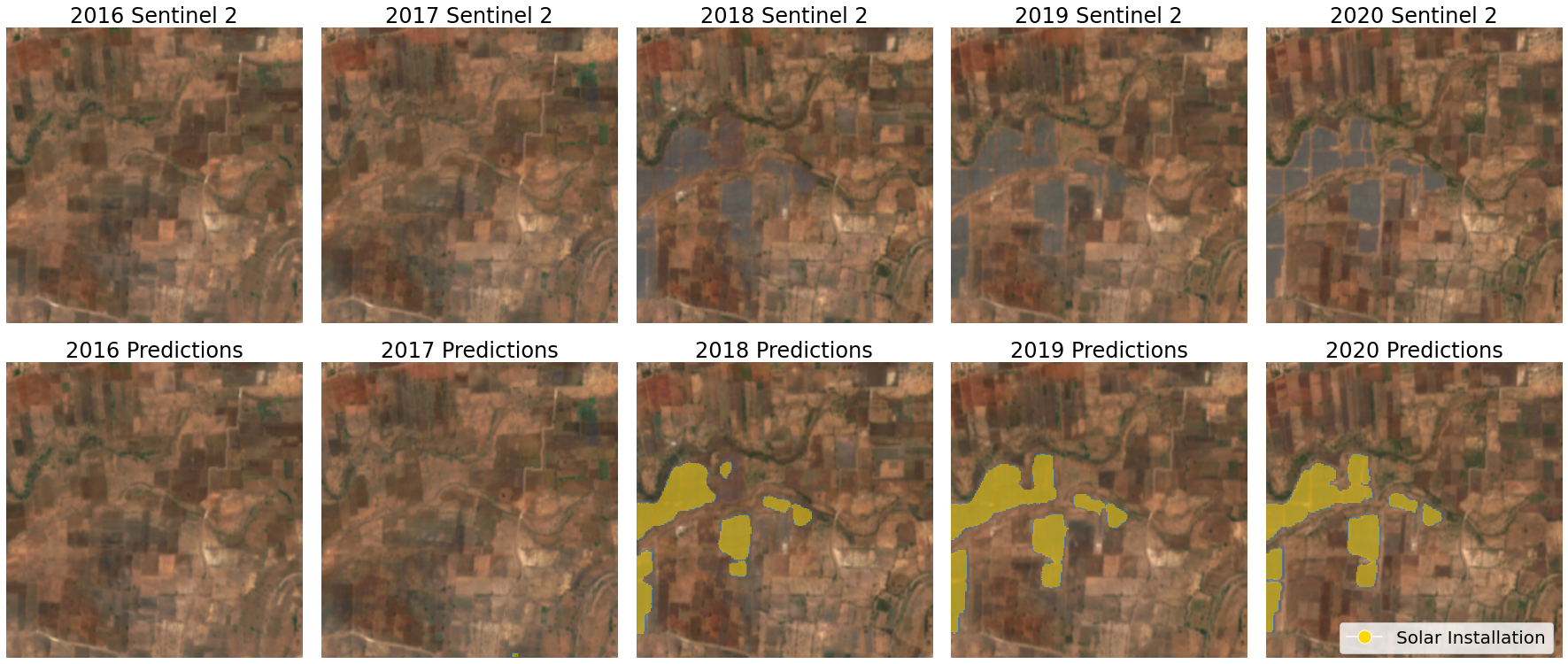}
\caption{Predictions over time. Solar farms predictions across time for sentinel imagery median composites for the years 2016, 2017, 2018, 2019, 2020. These predictions allow the Pearson correlation analysis between Karnataka solar install capacity and model predictions.}
\label{fig:solarchange}
\end{figure*}

To further study the performance of our model, we conducted a Pearson correlation coefficient analysis between installed solar capacity and the predicted total solar installation area for the state of Karnataka in India. To do this, we run inference for the different median composite Sentinel 2 imagery after histogram matching. Model predictions were polygonised and used to estimate the area of individual predictions. The total solar farm area predicted by our model is used to make a correlation analysis with the solar install capacity presented in Table \ref{table:install-capacity}. 

\begin{figure}[!th]
    \centering
    \includegraphics[width=.7\linewidth]{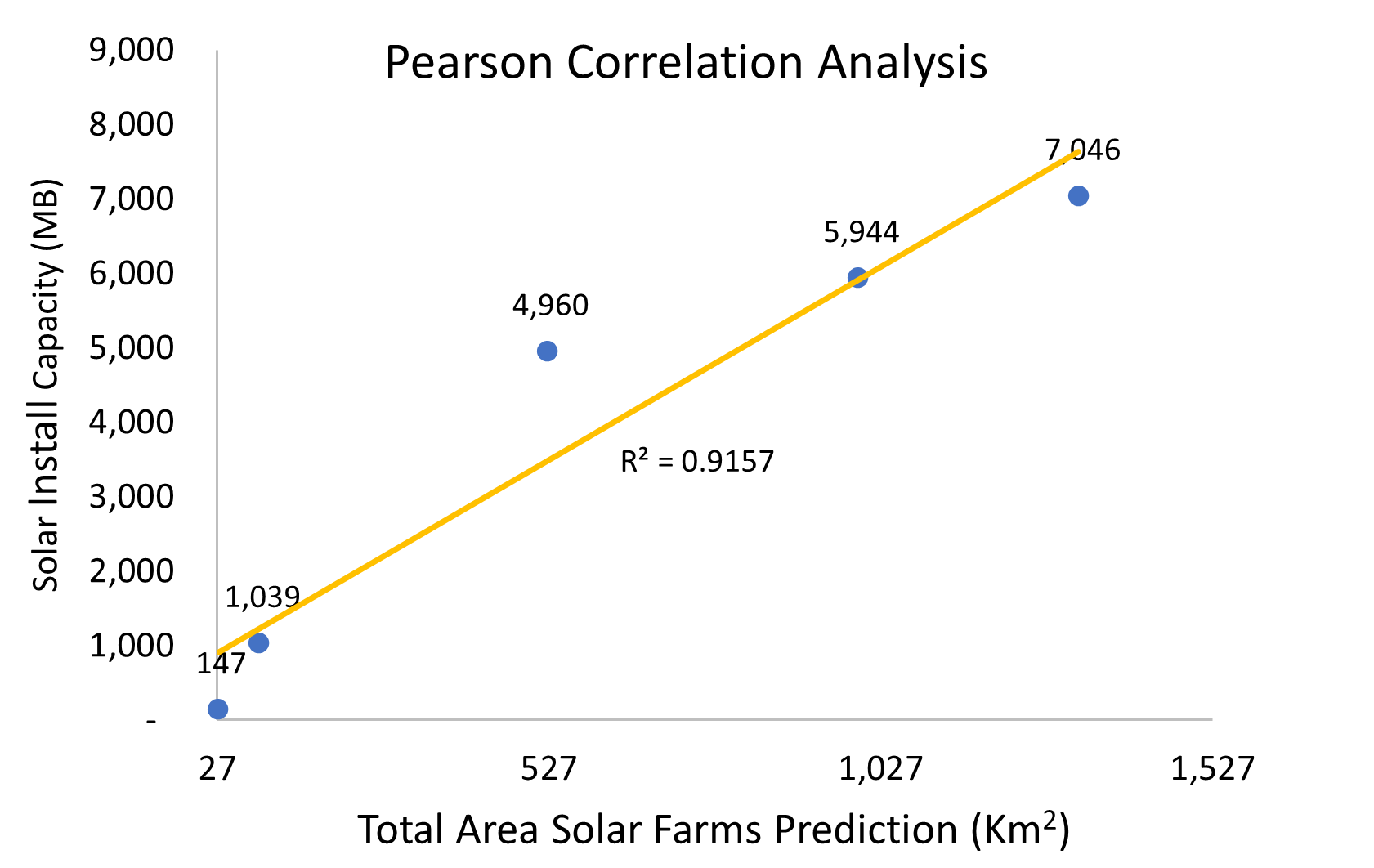}
    \caption{Pearson correlation between total area solar farms across time predicted by our model for the state of Karnataka and the solar install capacity in Thousand BTU's per hour units (MB)}
    \label{fig:solarcorrelation}
\end{figure}

Figure \ref{fig:solarcorrelation} shows the Pearson correlation between total area of solar farms across time predicted by our model for the state of Karnataka and the total installed solar capacity in Thousand BTU’s per Hour (MB). The Pearson correlation coefficient is $\textbf{0.957}$ indicating a very strong relationship between our model predictions and the solar install capacity released by the Indian state of Karnataka. The coefficient of determination ($R^{2}$) or proportion of the variance of solar install capacity explained by our model predictions is $\textbf{91.57\%}$.

\subsection*{Land Cover land Use change analysis results}

Table \ref{table:landcover-results} shows the percentage of each land cover class converted by solar PV installations across India. Over 74\% of the solar farms installations in India was built  on land cover types that could create potential biodiversity and food security conflicts - 67.6\% of agriculture land and 6.99\% of natural habitat- of which 38.6\% of agricultural land may have potential to cultivate seasonal crops including Kharif (Kharif crops, or monsoon crops are domesticated plants that are cultivated and harvested during the Indian subcontinent’s monsoon season), Rabi (Rabi crops are agricultural crops that are sown in winter and harvested in the spring), and Zaid (Zaid crops are summer season crops), and 28.95\% of land with plantation crop/orchards. The natural land cover types included sensitive ecosystems such as evergreen, deciduous, and littoral swamp forest with potential biodiversity value. However, our results are sensitive to data limitation. Because, as we strictly restricted our model threshold to reduce false positive solar areas, we were only able to map around 20\% of currently installed utility-scale solar projects across India. Therefore, our results and interpretation of land use of impact of PV installations can change as and when future studies are able to map entire utility scale solar projects across India.

\begin{table}[ht]
\caption{Solar Farms Landcover Change Analysis using NRSC Land Use Land Cover data}
\label{table:landcover-results}
\centering
\resizebox{.48\textwidth}{!}{
\begin{tabular}{ r l }
\toprule
\textbf{Previous Landcover Class } & \textbf{Landcover Percentage (\%)} \\ 
\midrule
 Build Up &  0.15\% \\ 
 Kharif Only &  12.18\% \\ 
 Rabi Only & 18.49\%  \\ 
 Zaid Only & 7.97\%  \\
 Double/Triple & 0.00\%  \\
 Current Fallow & 11.50\% \\
 Plantation/Orchard & 28.95\% \\
 Evergreen Forest & 3.98\% \\
 Deciduous Forest & 0.08\% \\
 Scrub/ Deg. Forest & 1.88\% \\
 Littoral Swamp & 1.05\% \\
 Wasteland & 0.68\% \\
 Scrubland & 12.71\% \\
\bottomrule
\end{tabular}}
\end{table}

\begin{table}[ht]
\caption{Global Performance Assessment Results}
\label{table:manual-validation}
\centering
\resizebox{.48\textwidth}{!}{
\begin{tabular}{ r l l }
\toprule
\textbf{Category } & \textbf{\# of Records} & \textbf{Perc. Contribution} \\ 
\midrule
 Valid Farms & 4421 & 85.27\% \\ 
 Roof Top Solar & 387 & 7.27\%  \\ 
 Invalid Farms &  377 & 7.46\% \\ 
 \midrule
 \textbf{Total} &  5185 & 100.00\% \\ 
 
\bottomrule
\end{tabular}}
\end{table}

\subsection*{Manual Data Validation}

\begin{figure}[!th]
    \centering
    \includegraphics[width=\linewidth]{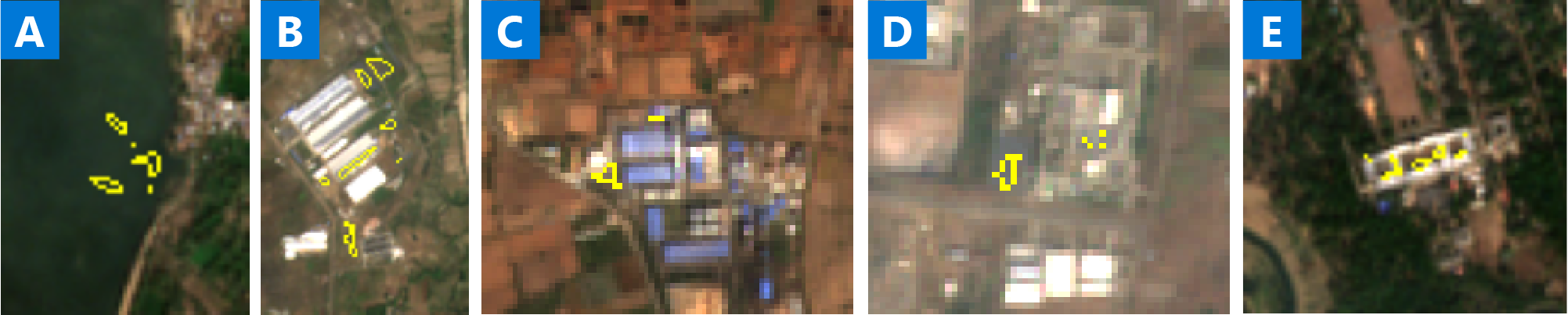}
    \caption{Examples of false positive predictions. The most common false positive predictions from our solar mapping model include electric plants (middle), construction sites (right), and seldom parking lots.}
    \label{fig:fp}
\end{figure}
To check the accuracy of our model predictions we performed a manual validation process. We overlaid our final model predictions after post-processing for the entire country of India on several base map layers inside QGIS and Google Earth software applications. We added Esri and Google Maps Satellite imagery in QGIS as a base map. We zoomed to each solar farm record and visually evaluated the overlay to tag the record either as a valid farm or invalid prediction or roof top (our model often mapped roof top solar farms as well). Figure \ref{fig:solarpredictions} show multiple example of valid solar farms predictions. Figure \ref{fig:fp} show examples of invalid (A, C) and roof top solar predictions (B, D, E). We also cross-checked this by using historical high resolution imagery available from Google Earth Pro’s “Show historical imagery” feature. In some cases where the use of satellite imagery was not conclusive enough we also conducted Internet searches to identify reports or news about presence of solar farms in a given area. For example, public reports where used to validate the solar farm footprint for Rewa Solar power plant in Madhya Pradesh since part of the solar footprint was not current in the basemaps used for reference.

\subsection*{Cross India Solar PV Farms Database generation}

\begin{figure}[!th]
\centering
\includegraphics[width=.8\linewidth]{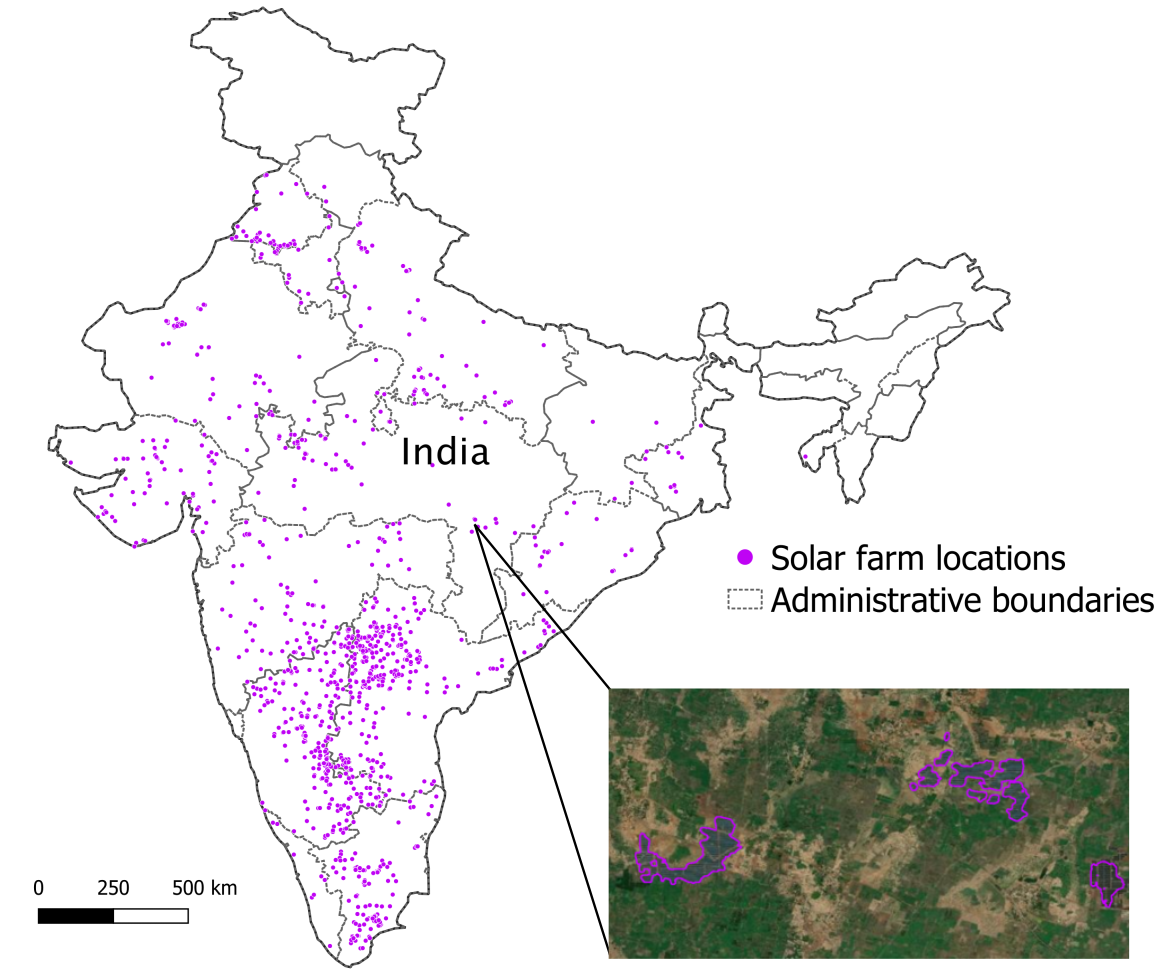}
\caption{Solar Farm Detection Results. Map showing center points of solar farms that were detected by our model. Note: The boundaries of States of India shown here are neither authenticated nor verified and are not to scale. They are only meant for graphical representation. All efforts have been made to make them accurate, however, neither Microsoft nor TNC own any responsibility for the correctness or authenticity of the same.}
\label{fig:database}
\end{figure}

 Most large scale solar PV farms include tens of  thousands of solar panels arranged in a non contiguous way. Hence, our model would predict multiple independent polygons. The 4421 manually validated correct predictions from the previous analysis were spatially cluster based on a distance metric into multi-polygons to obtain 1076 individual solar PV farms. Table \ref{table:manual-validation} shows the manual validation results. \textbf{92.54\%} of model predictions correspond to valid solar farms (85.27\%) or roof top solar (7.46\%) with only 7.46\% of the predictions corresponding to invalid farms. Figure \ref{fig:database} shows center point locations for all predicted and validated solar farms in India  across all states.

\section*{Usage Notes}

Solar energy is projected to be the major contributor to the renewable energy capacity addition in India and across the globe in the next couple of decades.  Rapid deployment of renewable energy is critical to avoid the disastrous impacts of climate change. Using the power of artificial intelligence, we have developed a spatially explicit semantic segmentation model using noisy pixel-wise labels and hard negative mining to map utility-scale solar projects across India with a mean identification accuracy of 92\%. Application of this model across the globe can help identify factors driving land suitability for solar projects and help public agencies plan better to facilitate solar energy development apart from helping track progress on solar energy developed. In addition, by mapping spatial patterns of solar development we can better understand land-use changes that may be driven by utility-scale projects. Empowering stakeholders which such information will catalyze rapid development of renewable energy while ensuring limited impacts to local communities and natural ecosystems in the process.

Different approaches have been previously proposed for automatic detection of PV arrays from very high-resolution satellite imagery using machine learning~\cite{davidson2014modeling,yu2018deepsolar,malof2016automatic}. These approaches often rely on high resolution aerial imagery that is only freely available in the United States (with 1 m/px spatial resolution) and dense labels that are expensive to collect. This work shows that solar farm detection is feasible with lower resolution (10 m) imagery that is freely available worldwide and low-cost point.
Other studies use geospatial variables including population demographics, housing characteristics to determine the variables that are predictive of photovoltaic (PV) energy adoption~\cite{davidson2014modeling}. Concurrently to our work, Dunnett et al. published a global dataset for windmill and solar farm locations~\cite{dunnett2020harmonised} relying on the solar farms/windmills being previously present in Open Street Map (OSM)~\cite{OpenStreetMap}. Unfortunately, the methods that rely on surveys, OSM, and surrogate predictive variables are limited in completeness and scale. For example, Dunnett et al. includes 328 valid solar PV installations across India. Our approach, on the other hand, is able to detect 1076 solar farms including 809 never mapped before on OSM.Also, concurrently to our work, Kruitwagen et al. published a global inventory of solar installations using satellite imagery predictions~\cite{kruitwagen2021global}. This study included 372 solar PV installations across India, while missing many installations. 

Six years ago, the international community finalized the Paris Agreement—an historic international climate change agreement—that included new commitments from all countries and outlined a set of rules for the global system over the coming years. The agreement sets out a system to track the progress of countries towards their targets—including principles defining “transparency and accountability” provisions. The dataset we have developed for India if expanded to other countries could be a simple and transparent mechanism to track progress on the deployment of solar energy that can help hold countries accountable to deliver on their climate targets.  

Land-use and land-cover change is a pervasive, accelerating, and impactful process. Land-use and land-cover change is driven by human actions, and, in many cases, it also drives changes that impact humans. Understanding these patterns is critical for formulating effective environmental policies and management strategies. Because our dataset allows for both the identification of the spatial location of new solar development as well as the timing of that development, the dataset can be used in conjunction with land-use change models to better understand patterns of future change. Given the large land footprint associated with solar and onshore wind energy development there is potential for renewable energy expansion to involve the clearing of natural lands or fragmenting wildlife habitat and converting fertile agriculture land~\cite{kareiva2017energy}. Our analysis of past land use change driven by solar development in India indicates almost 7\% of development occurred within habitats important both for biodiversity and carbon storage i.e. evergreen, deciduous, and littoral swamp forest.  In the face of climate change, which is likely to interact strongly with other stressors, biodiversity conservation and agriculture food security requires proactive adaptation strategies~\cite{mawdsley2009review}. Renewable energy’s potential benefits to biodiversity from climate change mitigation will only be realized if development can prevent impacts to remaining natural habitat. Maintaining intact natural habitats and maintaining or improving the connectivity of land for the movement of both individuals and ecological processes, may provide the best opportunity for species and ecological systems to adapt to changing climate~\cite{anderson2010conserving}. 

On the other hand, the increasing demand for implementing renewable energy projects could put arable agriculture land under pressure~\cite{kiesecker2020renewable}. Globally fertile arable land suitable for agriculture is limited owing to natural conditions and nature protection, and are threatened by processes like urbanization, demographic shift, and climate change~\cite{ketzer2020land}. Further, overtime the demand for land to implement renewable energy projects expected to impact productive agriculture lands. Therefore, the loss of productive agriculture land may lead to new dimensions of land use conflicts and provoke economic, ecological, political, and social conflict disruptions, and may encourage food-versus-energy controversy. Given that we observed that nearly two thirds of solar development was located in agricultural areas, avoiding conversion to productive agricultural land will be an important strategy for renewable energy deployment. Thus, guiding renewable energy development toward areas with lower conflict  be important. Understanding the factors associated with renewable energy development and predicting future expansion patterns will allow to proactively identify potential conflicts between renewable energy and other important land uses. The first step in this process is having access to data on the locations of solar installations that can be regularly updated.

\section*{Code availability}

The dataset will be made publicly available for researchers, conservationist, policy makers, and solar developers to further explore conservation and solar energy development relationships, help inform policy decisions and minimize solar development effects in ecosystems at:
\url{https://researchlabwuopendata.blob.core.windows.net/solar-farms/solar_farms_india_2021.geojson} 
Source code with our model architecture implementation, trained models accompany with instructions on how to use it is available on GitHub at: \url{https://github.com/microsoft/solar-farms-mapping}   for anyone to use under MIT open-source license.

\section*{Acknowledgments}
We thank MacArthur Foundation of India, the Microsoft AI for Good initiative, and The Cynthia and George Mitchell Foundation for Funding Support.

\newpage
\bibliography{mybibfile}



\section*{Author contributions statement}

Must include all authors, identified by initials, for example:
A.O., D.N, S.M., J.K., and J.L.F. conceived the experiment(s), A.O., S.M., and C.R. conducted the experiment(s), A.O., D.N., J.K, and J.L.F. analysed the results. P.B. and S.M. performed the manual technical validation of results. All authors reviewed the manuscript. 

\section*{Competing interests} 

Authors declare no competing interest.




\end{document}